\documentclass[cameraready]{Interspeech}

\title{Segregate, Refine, Integrate: Decomposing Multimodal Fusion for Sentiment Analysis}

\author[affiliation={1, 5}, orcid=0009-0000-8850-0306]{Alexios}{Filippakopoulos}
\author[affiliation={1}, orcid=0009-0001-0366-7420, equalcontribution]{Elias}{Kallioras}
\author[affiliation={1,2}, orcid=0009-0000-3952-693X, equalcontribution]{Nikolaos}{Xiros}
\renewcommand{\authorsep}{,\\}
\author[affiliation={1,3,\ddagger}, orcid=0000-0002-6042-9584, correspondingauthor]{Efthymios}{Georgiou}
\author[affiliation={1, 4, 5}, orcid=0009-0007-1532-5288]{Alexandros}{Potamianos}

\address{
    $^1$ National Technical University of Athens, Greece, $^2$ Athena Research Center, Greece, \\$^3$ University of Bern, Switzerland, $^4$ Archimedes AI, Greece, $^5$ Synaptic Bloom PBC, US
}

\email{alexiosfilippakopoulos@mail.ntua.gr, eliaskallioras@gmail.com, n.xiros@athenarc.gr, efthymios.georgiou@unibe.ch, potam@central.ntua.gr}


\usepackage{comment, svg, pifont}
\usepackage{multirow}
\usepackage{threeparttable}

\begin{document}

\maketitle

\ifcameraready
    {\def\thefootnote{$\ddagger$}\footnotetext{Work done while at the National Technical University of Athens.}}
\fi

\begin{abstract}
Multimodal fusion must simultaneously refine modality-specific signals and model cross-modal interactions; two competing objectives typically entangled within the same operation. We propose \textbf{SeRIn} (\textbf{Se}gregate, \textbf{R}efine, \textbf{In}tegrate), a multimodal LM fusion scheme that enforces this separation as an architectural prior. Modality-specific representations evolve along isolated pathways, each refined against its respective encoder context, while a dedicated cross-modal pathway accumulates their joint evolution without contaminating unimodal streams. Full cross-modal interaction is deferred to a final prediction step — ablations confirm that structured interactions, not added capacity, drive the gains; gate analysis under visual corruption reveals emergent modality reweighting without explicit supervision. SeRIn achieves state-of-the-art results on CH-SIMS and CMU-MOSEI, improving all metrics on both benchmarks.
\end{abstract}

\section{Introduction}
Multimodal sentiment analysis (MSA) aims to predict the polarity 
and intensity of speaker sentiment by jointly modeling linguistic 
content, acoustic signals, and visual cues. It is a key capability 
for affective computing applications including conversational agents, 
mental health assessment, and social media opinion mining~\cite{baltruvsaitis2018multimodal, liang_foundations_2023}. 
The central technical challenge, \emph{multimodal fusion}, involves learning 
representations that capture both unimodal cues and cross-modal interactions. This is non-trivial 
because the three modalities carry different types of information and reside in different geometries and timescales~\cite{georgiou2025multimodal}. Furthermore, their interactions can be subtle, acoustic and visual cues 
can reinforce, nuance, or outright contradict lexical content, as 
in sarcasm or emotional suppression~\cite{mosei, poria2017review, sims}.

Two objectives arise naturally in multimodal fusion: refining each modality and modeling their interactions. These objectives are typically entangled within a single fusion operation. Tensor
and attention-based methods~\cite{tfn, lmf, MulT, magbert, cenet} 
perform both operations simultaneously, with no architectural separation between the two. 
Disentanglement and auxiliary-objective approaches~\cite{misa, fdmer, selfmm, drtsc} encourage 
modality-specific representations, but place no structural restriction on multimodal fusion during the forward pass, specialization depends on learned 
penalties rather than architectural constraints. We draw inspiration from DeepMLF~\cite{deepmlf}, which introduced learnable fusion tokens within a frozen language model (LM), 
providing meaningful separation between the text stream and the multimodal pathway.
However, among the fusion tokens themselves, that act as multimodal information carriers, the 
model is left to implicitly decide when to preserve modality-specific signals and when to integrate them. 
More broadly, \emph{interaction topology}, the pattern specifying which representations may exchange 
information, under what conditions, and at which processing stage, remains an underexplored design axis in multimodal fusion, complementary to depth and capacity. In particular, we argue that unimodal refinement and cross-modal integration should be treated as distinct stages rather than being coupled within a single fusion operation.

We propose \textbf{SeRIn}\footnote{Code is available on GitHub at
\href{https://github.com/alexisfilippakopoulos/SeRIn-MSA}{SeRIn-MSA}.}
\footnote{Accepted at Interspeech 2026}
(\textbf{Se}gregate, \textbf{R}efine, \textbf{In}tegrate), which enforces this separation as an architectural prior rather than an optimization penalty. Built on the learnable fusion token design of DeepMLF~\cite{deepmlf}, SeRIn (i)~\textbf{Segregates} the fusion tokens into per-modality groups confined to isolated pathways, alongside a cross-modal pathway that reads and aggregates them but never writes back; (ii)~\textbf{Refines} each pathway against its own encoder context through internally gated attention; and (iii)~\textbf{Integrates} all representations under unrestricted cross-modal interaction only at the final prediction step. Segregation is enforced structurally, through modality-constrained attention masks that add no parameters, so that each modality is refined before any cross-modal mixing. With fusion depth and token count held fixed to DeepMLF's optima, interaction topology is the sole free variable, and our ablations attribute the gains to this structure, not to added capacity. Our contributions can be summarized as:
\begin{enumerate}

\item We frame interaction topology as a fusion design axis complementary to depth and capacity, and realize it in SeRIn, which segregates per-modality pathways via parameter-free masks, refines each against its own encoder context, lets an audiovisual pathway read-but-not-write, and defers unrestricted interaction to a final head.

\item SeRIn reaches SOTA on CH-SIMS and CMU-MOSEI across all metrics with depth and token count fixed; a capacity-matched ablation that keeps the added parameters but removes the topology falls below DeepMLF, and gate analysis under visual corruption reveals emergent, unsupervised modality reweighting

\end{enumerate}







\section{Related Work}

Standard MSA taxonomies (early/late/deep stage, fusion mechanism) capture integration timing but not the structural constraints governing how modality pathways interact during the forward pass. To motivate SeRIn, we reinterpret prior work along two axes: (i) fusion depth and interaction topology, and (ii) the mechanism of unimodal specialization: optimization objectives versus architectural constraints.

\subsection{From Shallow to Deep Interaction Modeling}

Early multimodal sentiment models perform cross-modal interaction in a single integration stage. Tensor Fusion Network (TFN)~\cite{tfn} and Low-rank Multimodal Fusion (LMF)~\cite{lmf} compute explicit cross-modal interactions but do not separate unimodal refinement from multimodal integration within the architecture.

Attention-based models introduce more flexible interaction patterns. MulT~\cite{MulT} models directional pairwise attention across every ordered modality pair, constraining the direction of influence across modalities. Cross-modal attention architectures such as CENet~\cite{cenet} and MAG-BERT~\cite{magbert} inject acoustic and visual context into textual representations through attention modules. Although these methods improve cross-modal expressivity, unimodal refinement and multimodal integration typically occur within the same attention operations, and cross-modal communication becomes unrestricted once introduced.

Frozen pretrained language models enable layer-wise fusion within a fixed backbone, where interleaving fusion modules with pretrained layers makes interaction structure an explicit design choice. Flamingo~\cite{flamingo} and Audio Flamingo~\cite{audio-flamingo} insert gated cross-attention modules between frozen decoder layers, using scalar residual gates to regulate the overall magnitude of injected multimodal signals. These gates apply scalar modulation, whereas SeRIn uses content-dependent, element-wise gating on the attention output before projection.
Deep fusion in MSA traces back to DHF~\cite{georgiou2019deep}, the first such paradigm, which learns unimodal encoders followed by a fusion network. DeepMLF~\cite{deepmlf} brings this lineage into the frozen-LM setting, demonstrating that fusion depth and multimodal capacity are critical factors and introducing learnable fusion tokens to mediate cross-modal interaction across layers. However, the fusion token pool is undifferentiated, allowing every fusion token to accumulate the same pre-fused joint audiovisual context. As a result, modality-specific structure is not explicitly enforced by the forward architecture but emerges, if at all, through optimization. SeRIn adopts DeepMLF, whose fusion token interface provides the controlled substrate needed to study interaction topology in isolation, a variable DeepMLF and most prior work leave unconstrained. Concretely, SeRIn partitions these tokens into modality-specific pathways with constrained interaction, adds a read-only audiovisual pathway that maintains a cross-modal state without modifying unimodal streams, and defers unrestricted cross-modal interaction to a final integration head.

\subsection{Objective-Based vs Architectural Specialization}

A parallel line of work promotes unimodal specialization through representation-level constraints. These approaches decompose each modality into shared (modality-invariant) and modality-private subspaces: MISA~\cite{misa} enforces the split with a soft-orthogonality loss, whereas FDMER~\cite{fdmer} learns the two subspaces adversarially through a modality discriminator. DRTSC~\cite{drtsc} extends this framework with temporal smoothness losses and adversarial alignment to encourage consistency while preserving modality-specific components. Other approaches steer specialization through training objectives, learned reweighting, or input re-representation. Self-MM~\cite{selfmm}, JTUM~\cite{jtum}, TETFN~\cite{tetfn}, and MTFN~\cite{mtfn} introduce auxiliary supervision or multi-task objectives to reinforce unimodal structure; KuDA~\cite{kuda} injects sentiment knowledge to dynamically reweight each modality's contribution per sample; and DEVA~\cite{deva} converts raw audio and visual signals into fine-grained textual emotional descriptions prior to a text-guided progressive fusion. In all these cases, however, the forward computational graph places no structural restriction on cross-modal mixing; modality-specific behavior is shaped by learned objectives or input re-representation rather than by architectural constraints.

Architectural specialization has been explored more recently in DLF~\cite{dlf}, which enforces asymmetric attention toward language (V$\rightarrow$L, A$\rightarrow$L, L$\rightarrow$L) as a structural constraint. While this controls the direction of influence, every interaction is funneled into the language stream — audio and visual features act only as sources and never interact with each other — and language self-refinement (L$\rightarrow$L) and cross-modal integration (V/A$\rightarrow$L) remain intertwined within each layer.

In contrast, SeRIn seals unimodal pathways by construction and restricts cross-modal interaction to a dedicated read-only pathway and a final integration stage, constraints DLF and prior work do not enforce. This architecturally decouples unimodal refinement from multimodal integration during the forward pass, rather than delegating the trade-off to optimization — to our knowledge, a distinct departure from existing MSA fusion strategies.


\section{Preliminaries}

We formalise the MSA task and establish the notation used
throughout the paper.

\subsection{Problem Formulation}

MSA is formulated as a supervised regression task that infers
sentiment polarity and intensity from three modalities:
text, audio, and visual.
Let $\mathcal{M} = \{t, a, v\}$ be the modality index set.
For each $m \in \mathcal{M}$, the input for sample $i$ is
$\mathbf{X}_{i,m} \in \mathbb{R}^{L_m \times D_m}$,
where $L_m$ is the maximum sequence length and $D_m$ the
feature dimensionality.
The goal is to learn
$f_\theta : \prod_{m \in \mathcal{M}}\mathcal{X}_m \to \mathbb{R}$
mapping each multimodal input
$\mathcal{X}_i = \{\mathbf{X}_{i,m}\}_{m \in \mathcal{M}}$
to a scalar sentiment label $y_i \in \mathbb{R}$.

\subsection{Transformer Backbone}
\label{transf-prel}

SeRIn is instantiated within a frozen pre-norm decoder-only
Transformer~\cite{attentionneed,decoder-only}.
Each LM Block applies causal self-attention (CSA)
followed by a feed-forward network:
\[
  \tilde{\mathbf{H}}^{(l)} = \mathbf{H}^{(l-1)} + \mathrm{SA}\bigl(\mathrm{Norm}(\mathbf{H}^{(l-1)}); \mathbf{M}^{\mathrm{causal}}\bigr)
\]
\[
  \mathbf{H}^{(l)} = \tilde{\mathbf{H}}^{(l)} + \mathrm{FFN}\bigl(\mathrm{Norm}(\tilde{\mathbf{H}}^{(l)})\bigr).
\]

The causal mask $\mathbf{M}^{\mathrm{causal}}$ restricts each
token to attending only to itself and preceding positions.
When fusion tokens are appended to the text sequence,
$\mathbf{M}^{\mathrm{causal}}$ is replaced by a
modality-constrained mask $\mathbf{M}^{\mathrm{mc}}_{\mathrm{LM}}$,
the structural mechanism through which SeRIn enforces modality
segregation inside the frozen LM (see Sec.~\ref{LM Block}).

\subsubsection{Attention masking}
\label{}
A binary mask $\mathbf{M} \in \{0,1\}^{n_q \times n_k}$
prevents selected query–key interactions by injecting $-\infty$
before the softmax:

\[
  \mathrm{Attn}(\mathbf{Q},\mathbf{K},\mathbf{V};\mathbf{M})
  = \sigma\left(
      \frac{\mathbf{Q}\mathbf{K}^{\top}}{\sqrt{d_k}}
      + (1-\mathbf{M})(-\infty)
    \right)\mathbf{V}.
\]

\begin{figure*}[t]
    \centering
    \includegraphics[width=\textwidth]{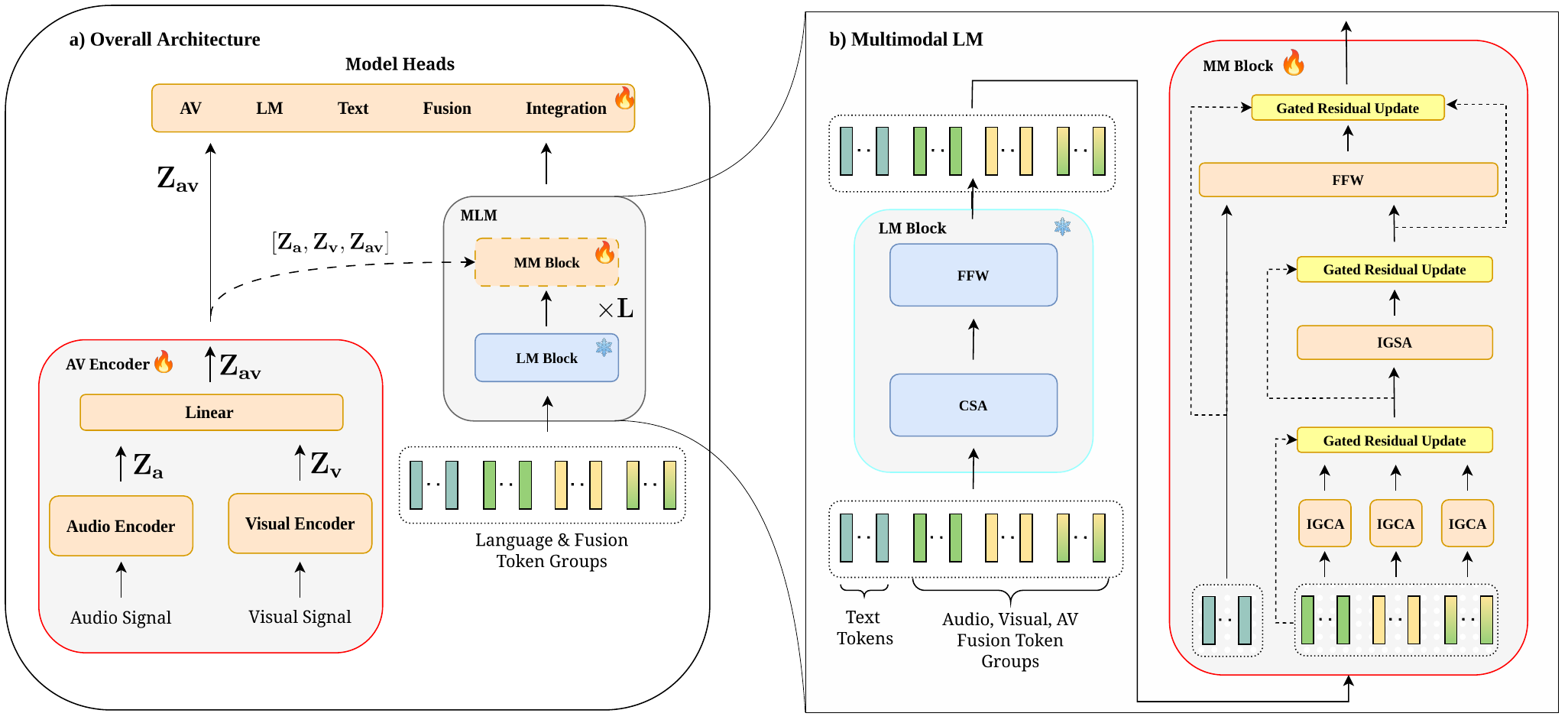}
        \caption{Overview of the SeRIn architecture. Audio and visual inputs are encoded by modality-specific encoders and injected into a frozen language model via modality-partitioned fusion tokens. Structured MM Blocks enforce segregation through constrained attention, regulate information flow via internal gating, and enable staged integration before final prediction by the Integration Head.}
    \label{fig:overall_architecture}
\end{figure*}

$M_{ij}=1$ permits query $i$ to attend to key $j$, while
$M_{ij}=0$ suppresses it. This formulation is used directly in
Secs.~\ref{LM Block} and~\ref{MM Block}
to define $\mathbf{M}^{\mathrm{mc}}_{\mathrm{LM}}$
and $\mathbf{M}^{\mathrm{mc}}_{\mathrm{MM}}$. The unimodal encoders
$\mathcal{E}_a, \mathcal{E}_v$ (\autoref{fig:overall_architecture})
are standard pre-norm Transformer encoders~\cite{attentionneed} and share
the same layer structure as described above, with
$\mathbf{M}^{\mathrm{causal}}$ replaced by an unrestricting (full) attention mask.

\section{Methodology}

\subsection{Overview}
\label{overview}
We present \textbf{SeRIn} (\textbf{Se}gregate, \textbf{R}efine, \textbf{In}tegrate), a multimodal fusion framework that injects audiovisual information into a frozen pretrained language model through structured, gated cross-modal pathways.
\begin{enumerate}
\item \textbf{Segregate}. The fusion process is organized around two structurally distinct representational pathways: modality-specific pathways, sealed from auxiliary-modality influence while remaining grounded in linguistic context, and a cross-modal pathway that progressively reads from them and aggregates their joint evolution. These pathways are realized through learnable fusion tokens partitioned into disjoint modality-specific groups, with one-way flow enforced via modality-constrained attention masks.
\item \textbf{Refine}. Within each pathway, learned gated modules actively update and consolidate representations without breaching segregation. Internally Gated Cross-Attention (IGCA) injects fresh encoder context into each pathway through content-dependent element-wise gates, while Internally Gated Self-Attention (IGSA) consolidates the injected information into a coherent pathway-level representation.
\item \textbf{Integrate} completes the staged design by lifting all segregation constraints only at prediction time. A standard late-fusion encoder mechanism aggregates the refined representations, ensuring unrestricted cross-modal interaction occurs solely where it is appropriate: the final prediction step.
\end{enumerate}
We instantiate SeRIn within the frozen-LM paradigm of DeepMLF~\cite{deepmlf}, whose fusion token interface provides a controlled substrate for studying how interaction topology shapes multimodal representation learning.

As illustrated in~\autoref{fig:overall_architecture}, the model consists of three components: (i)~an \textit{AV Encoder} that independently encodes audio and visual sequences into latent representations $\mathbf{Z}_a$, $\mathbf{Z}_v$, and $\mathbf{Z}_{av}$; (ii)~a \textit{Multimodal Language Model} comprising a frozen pretrained decoder augmented with modality-specific fusion tokens and MM Blocks that perform internally gated cross-attention and internally gated modality-constrained self-attention at selected layers; and (iii)~an \textit{Integration Head} that aggregates all representations for final sentiment prediction. The input sequence to the LM follows the layout:
\[
\mathbf{H}^0 = [\mathbf{X}_t^0;\mathbf{X}_{f_a}^0;\mathbf{X}_{f_v}^0;\mathbf{X}_{f_{av}}^0] \in \mathbb{R}^{(L + n_a + n_v + n_{av}) \times d}
\]
where $\mathbf{X}_t \in \mathbb{R}^{L_t \times d}$ denotes the text token embeddings and $\mathbf{X}_{f_m} \in \mathbb{R}^{n_m \times d}$ denotes the learnable fusion tokens for modality $m \in \{a, v, av\}$, with total fusion token count $n_f = n_a + n_v + n_{av}$.

\subsection{AV Encoder} The AV Encoder is a dual-stream Transformer-based architecture designed to learn temporally aligned representations from audio and visual sequences. Given the audio and visual input signals, the model first applies a deterministic alignment procedure to map each sequence to a fixed maximum length $L_{\text{enc}}$. This is achieved via padding, if necessary, followed by segment-wise mean pooling, ensuring consistent temporal resolution across modalities. Each modality is then projected into a shared latent dimension $d_{\text{enc}}$, after which sinusoidal positional embeddings are added to encode temporal order. The unimodal representations are processed independently by stacks of $N$ Transformer encoder layers $\mathcal{E}_a, \mathcal{E}_v$, each consisting of multi-head self-attention with $H$ heads, a position-wise feed-forward network, and pre-layer normalization. This produces the encoded unimodal sequences $Z_a, Z_v \in \mathbb{R}^{L_{\text{enc}} \times d_{\text{enc}}}$. To obtain a joint multimodal representation, the encoded features are concatenated along the feature dimension and passed through a linear fusion layer, yielding $Z_{av} \in \mathbb{R}^{L_{\text{enc}} \times d_{\text{enc}}}$. This encoder follows~\cite{deepmlf} and is kept fixed to isolate interaction topology.
\[
\mathbf{Z}_\text{a} = \mathcal{E}_a(\mathbf{X}_\text{a}) \quad \mathbf{Z}_\text{v} = \mathcal{E}_v({\mathbf{X}_\text{v}}) \quad \mathbf{Z}_{\text{av}} = \text{FFW}(\mathbf{Z}_\text{a} || \mathbf{Z}_\text{v})
\]

\subsection{Modality-Specific Fusion Tokens}
Prior work appends a single pool of undifferentiated learnable fusion tokens to the LM input~\cite{deepmlf}, providing no structural distinction between intra-modal and cross-modal interactions. SeRIn removes this ambiguity by initially partitioning the $n_f$ fusion tokens into three disjoint, modality-specific groups (audio, visual, audiovisual) based on their interaction role:
\[
\mathbf{X}_{f_m} \in \mathbb{R}^{n_m \times d}, \quad m \in \{\text{a}, \text{v}, \text{av}\}, \quad n_f = n_\text{a} + n_\text{v} + n_\text{av}.
\]
Each group $\mathbf{X}_{f_m}$ consists of learnable parameters initialized from $\mathcal{N}(0, 0.02)$ and appended to the text embedding sequence before being fed into the first LM layer. This partitioning is structurally enforced through (i) the constrained self-attention mask applied in both the LM and MM blocks, and (ii) the modality-specific cross-attention routing within each MM block.

The unimodal groups are strictly isolated, allowing audio and visual representations to evolve along independent pathways throughout fusion. All three fusion token groups attend to the full text sequence at every LM layer, ensuring continuous linguistic grounding as representations develop. Within this shared grounding, each unimodal group builds specialized representations along its dedicated pathway: accumulating modality-specific context within its isolated group at every LM layer and receiving targeted encoder injections through its IGCA module at each MM Block.

The audiovisual (AV) group plays a distinct role: its tokens form a dedicated cross-modal pathway that continuously reads and aggregates the evolving states of both unimodal groups, enriching this summary at each MM Block via its IGCA module and the joint encoder representation $\mathbf{Z}_{\text{av}}$. Critically, AV tokens read from but never write to unimodal pathways, an asymmetry enforced by the Segregate principle that prevents cross-modal contamination while allowing the AV tokens to accumulate a layer-wise history of cross-modal co-evolution that neither the Integration Head nor $\mathbf{Z}_{\text{av}}$ alone can provide.

\subsection{Multimodal Language Model}

\subsubsection{LM Block with Modality-Constrained Self-Attention}
\label{LM Block}
The backbone of SeRIn is a frozen pretrained decoder-only language model. Each LM Block consists of a standard causal self-attention layer followed by a feed-forward network (Sec.~\ref{transf-prel}). When fusion tokens are appended to the text sequence, the attention mask must be adapted. Prior work~\cite{deepmlf} keeps text tokens causally masked among themselves while allowing every fusion token to attend to all text positions, so that fusion tokens absorb linguistic context without altering the text stream. We retain these two properties but go further by structuring the inter-group interactions among fusion tokens themselves.

Our modality-constrained mask is the structural mechanism through which the Segregate principle (Sec.~\ref{overview}) is enforced inside the frozen LM. By partitioning fusion token attention into modality-specific blocks, it turns an operation already present at every layer, the LM's self-attention, into the primary carrier of interaction topology, without additional parameters or computation.

We therefore adopt a \textit{modality-constrained} mask $\mathbf{M}^{\text{mc}}_\mathrm{LM} \in \{0, 1\}^{L \times L}$ that retains the text-causal and fusion-to-text properties above but additionally partitions the fusion token attention into modality-specific blocks. The mask has the following structure (where $\mathbf{C}$ denotes a causal block, $\mathbf{1}$ a full-attention block, and $\mathbf{0}$ a blocked interaction):
\[
\mathbf{M}^{\mathrm{mc}}_\mathrm{LM} =
\begin{array}{c|cccc}
 & \mathbf{X}_\text{t} 
 & \mathbf{X}_{f_\text{a}} 
 & \mathbf{X}_{f_\text{v}} 
 & \mathbf{X}_{f_\text{av}} \\[3pt] \hline \\[-5pt]
\mathbf{X}_\text{t}      & \mathbf{C} & \mathbf{0} & \mathbf{0} & \mathbf{0} \\
\mathbf{X}_{f_\text{a}}  & \mathbf{1} & \mathbf{1} & \mathbf{0} & \mathbf{0} \\
\mathbf{X}_{f_\text{v}}  & \mathbf{1} & \mathbf{0} & \mathbf{1} & \mathbf{0} \\
\mathbf{X}_{f_\text{av}} & \mathbf{1} & \mathbf{1} & \mathbf{1} & \mathbf{1}
\end{array}
\]
\noindent Text tokens attend only to preceding text positions and never to fusion tokens, ensuring that the frozen LM's text representations remain identical to their pretrained behavior and are not altered by the multimodal pathway. This mask is precomputed and cached, introducing no additional parameters or computation.

\subsubsection{MM Block}
\label{MM Block}
The MM Block is the core fusion module of SeRIn, inserted at a selected subset of decoder layers $\mathcal{L} \subset \{0, \ldots, L_{\text{dec}} - 1\}$. At each such layer, the MM Block operates immediately after the frozen LM Block and applies three sequential gated stages to the fusion token representations. All LM parameters remain frozen, only the MM Block parameters are trained.

\vspace{4pt}
\noindent\textbf{Stage 1: Internally Gated Cross-Attention (IGCA).}
Each fusion token group independently attends to its designated encoder output through a dedicated IGCA module: unimodal groups query their respective unimodal encoder representations $\mathbf{Z}_\text{a}, \mathbf{Z}_\text{v}$, while the AV group queries the joint encoder representation $\mathbf{Z}_\text{av}$, enriching its cross-modal summary with a direct bottom-up audiovisual signal. The queries are derived from the (normalized) fusion tokens, and the keys and values from the AV Encoder representations:
\[
\mathbf{X}_\text{t}^{\ell}, \mathbf{X}_{f_\text{a}}^{\ell}, \mathbf{X}_{f_\text{v}}^{\ell}, \mathbf{X}_{f_{\text{av}}}^{\ell}  = \mathrm{split}(\mathbf{H}^{\ell})
\]
\[
\boldsymbol{\Delta}_{f_m}^{\ell} = \mathrm{IGCA}_m\left(\mathrm{Norm}(\mathbf{X}_{f_m}^{\ell}), \mathbf{Z}_m\right), \quad m \in \{a, v, av\}.
\]
Standard gated cross-attention~\cite{flamingo, audio-flamingo, deepmlf} controls information flow only at the residual-connection level, via a scalar gate on the output. Our \textit{Internally Gated Cross-Attention} (IGCA) places the gate \emph{inside} the attention mechanism, where the (normalized) fusion tokens query their respective AV encoder context $\mathbf{Z}_m$. A content-dependent gate $\mathbf{g} \in [0,1]^{n_m \times d}$ then modulates the attention output, element-wise \emph{before} the output projection; a gate placement that large-scale studies of standard attention have found most effective~\cite{qiu2025gated}:
\[
\mathbf{g} = \sigma(\mathbf{W}_g \, \mathbf{X}_q), \quad
\mathrm{IGCA}(\mathbf{X}_q, \mathbf{Z}_m) = \mathbf{W}_O\, (\mathbf{A} \odot \mathbf{g}),
\]
allowing the model to selectively suppress or amplify individual feature dimensions of the cross-attended representation on a per-token basis. The three cross-attention outputs are concatenated and added to the fusion tokens via a gated residual connection controlled by a learnable scalar $\alpha_1$:
\[
\mathbf{X}_f^{\ell} = [\mathbf{X}_\text{t}^{\ell}\hspace{2pt}\vert\vert\ \mathbf{X}_{f_\text{a}}^{\ell}\hspace{2pt}\vert\vert\ \mathbf{X}_{f_\text{v}}^{\ell}\hspace{2pt}\vert\vert\ \mathbf{X}_{f_{\text{av}}}^{\ell}]
\]
\[
\mathbf{\bar{X}}_f^{\ell} = \mathbf{X}_f^{\ell} 
+ \sigma(\alpha_1) \odot
\left[
\boldsymbol{\Delta}_{f_\text{a}}^{\ell} \,\vert\vert\, 
\boldsymbol{\Delta}_{f_\text{v}}^{\ell} \,\vert\vert\, 
\boldsymbol{\Delta}_{f_{\text{av}}}^{\ell}
\right].
\]
Since $\alpha_1$ is initialized to zero ($\sigma(\alpha_1) = 0.5$), the model can learn to close or open this pathway during training.


\vspace{4pt}
\noindent\textbf{Stage 2: Modality-Constrained Internally Gated Self-Attention (IGSA).}
After the IGCA stage, each fusion token carries a fresh injection of modality-specific context from its corresponding encoder stream. Before this propagates further, tokens must consolidate it: within each group, individually cross-attended views must be reconciled into a coherent group-level representation, while AV tokens must read from the updated unimodal groups to incorporate their most recently refined states into the evolving cross-modal summary. Unconstrained self-attention at this point would allow premature mixing of the just-injected unimodal signals, undermining the segregation that Stage~1 enforces. 

We therefore apply the same block-structured interaction pattern as $\mathbf{M}^{\mathrm{mc}}_\mathrm{LM}$ (Sec.~\ref{LM Block}), now instantiated as a dedicated learned attention layer over fusion tokens only, via mask $\mathbf{M}^{\mathrm{mc}}_\mathrm{MM} \in \{0,1\}^{n_f \times n_f}$. Thus, unimodal fusion tokens can attend to their own group, whereas the AV tokens can attend to all fusion tokens. The update is:
\[
\mathbf{\hat{X}}^{\ell}_f
=
\mathbf{\bar{X}}^{\ell}_f
+
\sigma(\alpha_2) \odot \,
\operatorname{IGSA}
\bigl(
\mathrm{Norm}(\mathbf{\bar{X}}^{\ell}_f);\,
\mathbf{M}^{\mathrm{mc}}_\mathrm{MM}
\bigr).
\]
IGSA shares the IGCA gating architecture but operates as masked self-attention over fusion tokens only, the mask $\mathbf{M}^{\mathrm{mc}}_\mathrm{MM}$ restricts the attention pattern as described above, and an element-wise sigmoid gate modulates the attention output before projection. The residual update is again controlled by a learnable scalar $\alpha_2$.

\vspace{4pt}
\noindent\textbf{Stage 3: Gated Feed-Forward Network.}
The final stage reassembles the full sequence by concatenating the (unmodified) text representations from the LM Block output with the updated fusion tokens, and applies a position-wise feed-forward network (FFN) with a gated residual over the joint sequence:
\[
\mathbf{\bar{H}}^{\ell} = [\mathbf{X}_\text{t}^{\ell}\hspace{2pt}\vert\vert\ \mathbf{\hat{X}}_{f_\text{a}}^{\ell}\hspace{2pt}\vert\vert\ \mathbf{\hat{X}}_{f_\text{v}}^{\ell}\hspace{2pt}\vert\vert\ \mathbf{\hat{  X}}_{f_{\text{av}}}^{\ell}]
\]
\[
\mathbf{{H}}^{\ell} = \mathbf{\bar{H}}^{\ell} + \sigma(\alpha_3) \odot \mathrm{FFN}\left(\mathrm{Norm}(\mathbf{\bar{H}}^{\ell})\right).
\]
To provide the FFN with a strong initialization, as in~\cite{deepmlf}, its weights are copied from the corresponding frozen LM layer's feed-forward network at model construction time and are finetuned during training. The residual update is controlled by a learnable scalar $\alpha_3$.

\subsubsection{Integration Head}

The Integration Head is architecturally conventional: a standard 
late-fusion Transformer encoder over pooled and projected modality 
summaries. Its role within SeRIn is nonetheless structurally precise. 
It is the sole site where the segregation enforced throughout 
representation learning is deliberately lifted, permitting 
unrestricted cross-modal interaction exclusively for final 
prediction. This staged deferral is not incidental, it is what 
makes the earlier segregation coherent.

By this point, the structural constraints enforced throughout have produced a set of complementary, specialized representations. The Integration Head aggregates all of these without restriction. Concretely, seven summary vectors are extracted via mean-pooling or last-token selection from each representation stream: the last valid text hidden state $\mathbf{h}_t^{(L)} \in \mathbb{R}^{d}$, mean-pooled fusion tokens $\bar{\mathbf{X}}_{f_m}^{(L)} \in \mathbb{R}^{d}$ for each $m \in \{\text{a,v,av}\}$, and mean-pooled AV Encoder outputs $\bar{\mathbf{Z}}_m \in \mathbb{R}^{d_{\text{enc}}}$ for each $m \in \{\text{a,v,av}\}$. Each is projected into a shared dimension $d_{\text{fuse}}$ via a learned linear mapping $\pi_k$:
\[
\mathbf{t}_k = \pi_k(\mathbf{s}_k) \in \mathbb{R}^{d_{\text{fuse}}}, 
\quad k = 1, \ldots, 7.
\]
A learnable \texttt{[CLS]} token is prepended, yielding 
$\mathbf{H}_{\text{fuse}} \in \mathbb{R}^{8 \times d_{\text{fuse}}}$, 
which is processed by a small Transformer encoder with full 
self-attention, the only attention operation in SeRIn without a 
masking constraint. A residual connection bypasses the encoder, 
and the \texttt{[CLS]} output is passed through post-normalization
and a linear head to produce the final prediction:
\[
\hat{y} = \mathbf{W}_{\text{task}} \cdot 
\mathrm{Norm}\Big(\mathcal{E}_{\text{fuse}}(\mathbf{H}_{\text{fuse}}) 
+ \mathbf{H}_{\text{fuse}}\Big)_{[0]}.
\]

\subsubsection{Auxiliary Prediction Heads}
\label{aux_heads}
In addition to the primary prediction from the Integration Head, SeRIn employs auxiliary linear heads that produce independent sentiment estimates from individual representation streams:
\begin{itemize}
    \item $\hat{y}_t = \mathbf{W}_t \, \mathbf{h}_t^{(L)}$: prediction from the last valid text hidden state.
    \item $\hat{y}_\text{av} = \mathbf{W}_\text{av} \, \bar{\mathbf{Z}}_\text{av}$: prediction from the mean-pooled AV Encoder output.
    \item $\hat{y}_{f_m} = \mathbf{W}_{f_m}\, \mathbf{X}_{f_m}^{(L)}$: per-token predictions from each fusion token group, for $m \in \{\text{a, v, av}\}$.
\end{itemize}
These heads serve as auxiliary objectives during training and are discarded at inference time. Their role is to provide direct gradient signal to specific model components: the text head encourages the frozen LM to produce sentiment-informative text representations, the AV head supervises the AV Encoder, and the per-modality fusion token heads ensure that each group, unimodal and cross-modal alike, absorbs task-relevant information from its designated source throughout the segregated refinement phase.

\begin{table*}[!t]
\centering
\caption{Performance comparison on CH-SIMS and CMU-MOSEI. SeRIn results are presented as averages across five independent runs. $\uparrow$ / $\downarrow$ higher or lower is better. $\dagger$: results from M-SENA~\cite{mao2022msena}, *: reproduced results, –: absent metrics in the original paper.}
\label{tab:main_results}
\begin{threeparttable}

\scriptsize  

\setlength{\tabcolsep}{8pt}  
\begin{tabular}{lcccccc|cccccc}

\toprule

\multirow{2}{*}{\textbf{Method}} 
& \multicolumn{6}{c}{\textbf{CH - SIMS}} 
& \multicolumn{6}{c}{\textbf{CMU - MOSEI}} \\

\cmidrule(lr){2-7}
\cmidrule(lr){8-13}

& Acc2 $\uparrow$ & F1 $\uparrow$ & MAE $\downarrow$ & Corr $\uparrow$ & Acc3 $\uparrow$ & Acc5 $\uparrow$
& Acc2 $\uparrow$ & F1 $\uparrow$ & MAE $\downarrow$ & Corr $\uparrow$ & Acc5 $\uparrow$ & Acc7 $\uparrow$ \\

\midrule

MulT$\dagger$
& 78.56 & 79.66 & 0.453 & 0.564 & 64.77 & 37.94
& 84.63 & 84.52 & 0.559 & 0.733 & 54.18 & 52.84 \\

Self-MM$\dagger$
& 80.04 & 80.44 & 0.425 & 0.595 & 65.47 & 41.53
& 85.15 & 84.90 & 0.531 & 0.765 & 55.53 & 53.87 \\

TETFN$\dagger$
& 81.18 & 80.24 & 0.420 & 0.577 & 63.24 & 41.79
& 86.21 & 86.11 & 0.537 & 0.770 & 55.78 & 53.90 \\

CENet$\dagger$
& 77.90 & 77.53 & 0.471 & 0.540 & 62.58 & 33.92
& 86.38 & 86.32 & 0.526 & 0.778 & 56.15 & 54.26 \\

JTUM
& -- & -- & -- & -- & -- & --
& 85.58 & 85.44 & 0.548 & 0.769 & -- & 53.29 \\

DLF
& -- & -- & -- & -- & -- & --
& 85.42 & 85.27 & 0.536 & 0.764 & 55.70 & 53.90 \\

DEVA
& 79.64 & 80.32 & 0.424 & 0.583 & 65.42 & 43.07
& 86.13 & 86.21 & 0.541 & 0.769 & 55.32 & 52.26 \\

DRTSC
& -- & -- & -- & -- & -- & --
& 86.40 & 86.50 & 0.531 & 0.776 & -- & 53.70 \\

KuDA
& 80.74 & 80.71 & 0.408 & 0.613 & 66.52 & 43.54 
& 86.46 & 86.59 & 0.529 & 0.776 & -- & 52.89 \\

MTFN 
& 81.56 & 81.27 & 0.423 & 0.583 & 67.77 & \underline{45.20} 
& 86.60 & 85.80 & 0.535 & 0.760 & -- & 54.50 \\

DeepMLF* 
& \underline{82.58} & \underline{82.77} & \underline{0.363} & \underline{0.713} & \underline{69.32} & 44.29 
& \underline{86.77} & \underline{86.77} & \underline{0.505} & \underline{0.800} & \underline{57.48} & \underline{55.57} \\

\midrule

\textbf{SeRIn} 
& \textbf{84.30}  & \textbf{84.42}   & \textbf{0.357}   & \textbf{0.732}   & \textbf{71.77}   & \textbf{48.32}  
& \textbf{87.79} & \textbf{87.78} & \textbf{0.493} & \textbf{0.810} & \textbf{58.44} & \textbf{56.52} \\

\bottomrule

\end{tabular}

\end{threeparttable}
\end{table*}

\subsection{Training Recipe}
\label{train-recipe}

\noindent\textbf{AV Encoder Pretraining.} The AV Encoder is pretrained independently on the target dataset using a linear head from $Z_{\text{av}}$, and then fine-tuned jointly with the fusion modules while the LM remains frozen. This follows~\cite{deepmlf}, where encoder initialization was shown to improve stability and performance, allowing a controlled comparison focused on interaction topology.

\noindent\textbf{MSA Loss}: Our primary objective is the following based on the Integration Head prediction.
\[
\mathcal{L}_\text{MSA} = ||y- \hat{y}||
\]

\noindent\textbf{Auxiliary Losses}: Auxiliary linear heads produce independent predictions from the text hidden state $h_t^{(L)}$, the AV Encoder output $\mathbf{Z}_{\text{av}}$, and each fusion token group $X_{f_m}$ (Sec.~\ref{aux_heads})
\[
\mathcal{L}_\text{aux} = \lambda_{\text{av}} \hspace{2pt} ||y- \hat{y}_{\text{av}}|| +
\lambda_{\text{t}} \hspace{2pt} ||y- \hat{y}_{\text{t}}|| +
\sum_{m\in\mathcal{M}} \lambda_{f_m} \hspace{2pt} ||y- \hat{y}_{f_m}||
\]
\noindent\textbf{Multimodal LM Loss:}
To reduce overfitting and preserve linguistic consistency, we employ a multimodal language modeling objective~\cite{embarrassingly, m3, GKOUMAS2021184}. This loss predicts each token conditioned on preceding tokens and the multimodal contexts $\mathbf{Z}_{\text{a}}, \mathbf{Z}_{\text{v}}, \mathbf{Z}_{\text{av}}$ from the AV Encoder:
\[
\mathcal{L}_\text{LM} = - \sum_{t=1}^{L} \log p_\text{LM} \big( x_t \mid x_{<t}, Z_{\text{a}}, Z_{\text{v}}, Z_{\text{av}} \big)
\]
This objective is also adopted in DeepMLF~\cite{deepmlf}, where ablations confirm its significance, preventing catastrophic forgetting of the pretrained LM. Simultaneously, we regularize the pretrained language embeddings to mitigate lexical dominance, a known failure mode in MSA where strong linguistic priors suppress non-linguistic signals. After tokenization, SeqAug~\cite{seqaug} is applied to the embeddings, combining augmentation with the LM objective to improve generalization, as in~\cite{deepmlf}.

\section{Experiments}
\subsection{Experimental Setting}
\subsubsection{Evaluation Datasets \& Multimodal Features}
\noindent\textbf{CMU-MOSEI}~\cite{mosei} is an English MSA benchmark of 23,453 utterance-level clips ($\approx66$h) from 1,000+ speakers across $\approx250$ topics, annotated on a seven-point Likert scale ($-3$ to $+3$). Text dominates audio and visual features by 21.98\% relative, with audio and visual contributing comparably~\cite{deepmlf}. Audio features are extracted via COVAREP~\cite{covarep} (74-dim: F0, MFCCs, HNR, glottal source, formants); visual via Facet (35-dim: action units, head pose, landmarks).

\noindent\textbf{CH-SIMS}~\cite{sims} is a Chinese MSA benchmark of 2,281 monologue utterances ($\approx2.3$h) from 60 films, TV dramas, and variety clips, with continuous sentiment scores ($-1$ to $+1$) from native annotators. Modality contributions are nearly balanced (3.72\% relative gap)~\cite{deepmlf}, making it well-suited for evaluating fusion strategies. Audio features are extracted via LibROSA~\cite{librosa} (33-dim: log-scaled F0, 20 MFCCs, 12 CQT coefficients); visual via OpenFace~\cite{openface} (709-dim: landmarks, gaze, head pose, intensity-coded action units).

\vspace{-4pt}
\subsubsection{Evaluation Metrics}\vspace{-2pt}
For multimodal sentiment analysis, the task is framed as a continuous prediction problem, evaluating performance using mean absolute error and Pearson correlation. Following standard practices, continuous predictions are also converted to categorical labels for classification, with datasets supporting binary or multi-class (three, five, or seven levels) sentiment schemes. We report relevant metrics, including accuracy and F1 scores, as appropriate.

\subsubsection{Implementation Details}
We implement SeRIn within the M-SENA~\cite{mao2022msena} framework, under which most baseline comparisons are also conducted. We use the AdamW optimizer~\cite{adamw} with $\beta_1=0.9$, $\beta_2=0.95$, a batch size of 32, and a one-epoch cosine annealed warmup. The learning rate is set to $1\times10^{-4}$ and all auxiliary and language modeling loss weights are set to 1.0 for both datasets. For MOSEI we use GPT2-large~\cite{gpt2} as the language model backbone, and for SIMS we use a Chinese GPT2-base~\cite{gpt-chinese}, reflecting the difference in dataset language and scale. To ensure a controlled comparison, fusion depth $\mathcal{L}$ and total 
fusion token count $n_f$ are kept fixed to their optima reported in~\cite{deepmlf}. For SIMS, $\mathcal{L} = \{5,\dots,11\}$ and $n_f=16$; for MOSEI, $\mathcal{L} = \{7, 14, 21, 28, 35\}$ and $n_f=12$. The fusion tokens are partitioned across modality-specific  groups as $n_a=n_v=7,\ n_{av}=8$ for SIMS and  $n_a=n_v=n_{av}=4$ for MOSEI. All experiments are conducted on a single NVIDIA A100 GPU.

\subsection{Computational Analysis}
\label{comp-anal}
SeRIn introduces additional parameter and compute overhead relative to DeepMLF, primarily from the Segregate principle: instantiating three separate IGCA modules per MM Block rather than one. On SIMS (GPT2-base), SeRIn requires 290 GFLOPs, 92M parameters, and an average inference time of $2.09\pm0.64$ ms/sample (vs. 243 GFLOPs/42M for DeepMLF), a $2.2\times$ parameter increase driven by the small GPT2-base backbone, where trainable MM Block and Integration Head parameters constitute a proportionally larger share of the total. On MOSEI (GPT2-large), the frozen LM dominates the budget and the gap narrows substantially: SeRIn reaches $3.00\pm0.76$ ms/sample with 1702 GFLOPs and 183M parameters, compared to 1682 GFLOPs and 83M for DeepMLF, a marginal 1.2\% compute overhead. Critically, all additional parameters are confined to the trainable MM Blocks and Integration Head. As demonstrated in Sec.~\ref{sub-ind-ablation}, Step 6 retains nearly all of these added parameters while removing only the structured interaction topology, and performance substantially degrades on both datasets, falling below the DeepMLF baseline despite the larger parameter count. This confirms that the parameter increase is a necessary but not sufficient condition for the observed gains, and that interaction topology is the operative factor.

\subsection{Comparison with Other Methods}
    \autoref{tab:main_results} compares SeRIn against state-of-the-art MSA methods spanning tensor fusion~\cite{tfn,lmf}, directional cross-modal attention~\cite{MulT,cenet}, representation disentanglement~\cite{misa,fdmer, drtsc}, auxiliary supervision~\cite{selfmm,tetfn,mtfn,jtum}, knowledge-guided reweighting~\cite{kuda}, textual emotion description~\cite{deva}, asymmetric language-focused attention~\cite{dlf}, and deep frozen-LM fusion~\cite{deepmlf}. SeRIn achieves state-of-the-art performance across both benchmarks on every reported metric. SeRIn improves all metrics: on SIMS by +1.72 Acc2, +1.65 F1, $-$0.6\% MAE, +1.9\% Corr, +2.45 Acc3, +3.12 Acc5; and on MOSEI by +1.02 Acc2, +1.01 F1, $-$1.2\% MAE, +1\% Corr, +0.96 Acc5, +0.95 Acc7. Ablations (Sec.~\ref{ablations}) confirm that interaction  topology, not added capacity, primarily drives these gains.

\subsection{Ablation Study}
\label{ablations}
All ablation experiments are conducted on both CH-SIMS and CMU-MOSEI, with results reported as averages over five independent runs.
\subsubsection{Subtractive and Independent Component Removal}
\label{sub-ind-ablation}
Tables~\ref{tab:subtractive_ablation} and~\ref{tab:indp_ablation} report sequential and independent component removal ablations respectively. In the sequential ablation, each component is removed from the model produced by the previous step; in the independent ablation, each component is removed from the full SeRIn model in isolation.
\begin{table}[!h]
\centering
\caption{Subtractive removal ablation. Each step removes one component from the model resulting from the previous step.}
\label{tab:subtractive_ablation}
\begin{threeparttable}
\footnotesize
\setlength{\tabcolsep}{10pt}
\begin{tabular}{l cccc}
\toprule

\multirow{2}{*}{Sub. Removal} &
\multicolumn{2}{c}{SIMS} &
\multicolumn{2}{c}{MOSEI} \\

\cmidrule(lr){2-3}
\cmidrule(lr){4-5}

& Acc2 $\uparrow$ & F1 $\uparrow$ & Acc2 $\uparrow$ & F1 $\uparrow$ \\

\midrule

0. None (SeRIn) & \textbf{84.30} & \textbf{84.42} & \textbf{87.79} & \textbf{87.78} \\
1. Integr. Head  & 83.70 & 83.79 & 87.01 & 87.04 \\
2. IGSA gate         & 82.61 & 82.92 & 86.65 & 86.69 \\
3. IGCA gate         & 82.06 & 82.17 & 86.59 & 86.56 \\
4. $\mathrm{M}_{\text{MM}}^\text{mc}$ & 81.18 & 81.42 & 86.07 & 86.03 \\
5. IGSA              & 82.38 & 82.45 & 86.51 & 86.54 \\
6. $\mathrm{M}_{\text{LM}}^\text{mc}$ & 81.07 & 81.34 & 85.47 & 85.50 \\

\bottomrule
\end{tabular}
\end{threeparttable}
\end{table}

The sequential trajectory is broadly monotonically decreasing across both datasets, confirming that all components contribute meaningfully within the integrated framework. When the Integration Head is removed, it is replaced with a concatenation followed by a two-stage MLP, an implicit ablation of its Transformer encoder against a simpler aggregation baseline. The consistent drop across both tables ($-0.60$ Acc2, $-0.63$ F1 on SIMS; $-0.78$ Acc2, $-0.74$ F1 on MOSEI) confirms that unrestricted cross-modal interaction at prediction time, not merely the aggregation of the seven summary vectors, drives its contribution. The most structurally informative pattern in Table~\ref{tab:subtractive_ablation} spans Steps 4 through 6 and replicates across both benchmarks. Removing the MM-level mask (Step 4) produces a pronounced drop on both datasets, but subsequently removing IGSA (Step 5) partially recovers performance. This non-monotonicity is a direct mechanistic signature of the Segregate principle: without the modality-constrained mask, IGSA operates as unconstrained self-attention over all fusion tokens, instantiating precisely the premature cross-modal mixing the framework is designed to prevent. Its removal therefore reduces interference rather than eliminating a useful operation, demonstrating that IGSA and the MM-level mask are architecturally coupled — IGSA derives its value entirely from the structural context the mask provides. Table~\ref{tab:indp_ablation} confirms this directly: removing IGSA alone, with $\mathrm{M}_{\text{MM}}^\text{mc}$ intact, produces a clean drop on both datasets ($-1.02$ SIMS, $-0.89$ MOSEI), the opposite sign from the sequential result. This sign reversal serves as clear evidence that, in our experimental setting, a component's contribution is fundamentally tied to the topology it operates within.

\begin{table}[!h]
\centering
\caption{Independent removal ablation. Each component is removed while retaining all others.}
\label{tab:indp_ablation}
\begin{threeparttable}
\footnotesize
\setlength{\tabcolsep}{10pt}
\begin{tabular}{l cccc}
\toprule
\multirow{2}{*}{Ind. Removal} &
\multicolumn{2}{c}{SIMS} &
\multicolumn{2}{c}{MOSEI} \\
\cmidrule(lr){2-3}
\cmidrule(lr){4-5}
& Acc2 $\uparrow$ & F1 $\uparrow$ & Acc2 $\uparrow$ & F1 $\uparrow$ \\
\midrule
None (SeRIn) & \textbf{84.30} & \textbf{84.42} & \textbf{87.79} & \textbf{87.78} \\
Integr. Head  & 83.70 & 83.79 & 87.01 & 87.04 \\
IGSA gate         & 83.59 & 83.65 & 87.41 & 87.43 \\
IGCA gate         & 83.98 & 84.12 & 87.50 & 87.50 \\
$\mathrm{M}_{\text{MM}}^\text{mc}$ & 82.45 & 82.44 & 86.22 & 86.23 \\
IGSA              & 83.28 & 83.34 & 86.9 & 86.91 \\
$\mathrm{M}_{\text{LM}}^\text{mc}$ & 82.08 & 82.10 & 85.98 & 86.01 \\
\bottomrule
\end{tabular}
\end{threeparttable}
\end{table}

Removing the LM-level mask yields the lowest performance across all metrics on both datasets in both Tables~\ref{tab:subtractive_ablation} and~\ref{tab:indp_ablation}; this establishes modality-constrained attention within the frozen LM backbone as a foundational contributor and suggests that interaction topology is particularly beneficial when enforced from the earliest stages of representation learning. The consistently larger drops on SIMS across every row of both tables reflect its balanced modality contributions: structural mechanisms protecting non-linguistic signals have greater impact when those signals carry more weight. The gate ablations in Table~\ref{tab:indp_ablation} should be interpreted carefully: removing either gate leaves the underlying operation intact — cross-attention or self-attention respectively — with only the gated residual update remaining. The independent drops therefore measure the marginal value of content-dependent element-wise modulation over the residual alone, not the value of the operation itself. That these gaps are modest yet consistent is expected: the gates' primary role is fine-grained flow control within pathways that the surrounding segregation topology has already structured. Finally, Step 6 of Table~\ref{tab:subtractive_ablation} constitutes an effective capacity-matched ablation, retaining nearly all added parameters — modality-specific pathways, dedicated IGCA modules — while removing only the structured inter-pathway interactions enforced by $\mathrm{M}^{\text{mc}}_{\text{LM}}$. Its marked underperformance on both SIMS ($-3.23$ Acc2, $-3.08$ F1) and MOSEI ($-2.32$ Acc2, $-2.28$ F1), falling below the DeepMLF baseline despite retaining more parameters, demonstrates that the added modules are necessary but not sufficient: their value is contingent on the structural context the masks provide, directly addressing the parameter gap noted in Sec.~\ref{comp-anal}, and consistent with prior DeepMLF ablations showing that increasing fusion token count or depth within these configurations does not improve performance~\cite{deepmlf}. Hence, we deduce that improvements stem from imposed interaction topology, not added capacity.

\subsubsection{Segregation Principle}
The Segregate principle rests on two structural claims: unimodal pathways must be protected from cross-modal contamination during representation learning, and AV tokens require asymmetric read access to both unimodal groups in order to aggregate their evolving states. We ablate these by varying the masking strategy at both the LM and MM levels across three conditions. \textbf{Constrained} (proposed): unimodal tokens are mutually isolated, while AV tokens attend to all groups. \textbf{Symmetric}: all groups are strictly isolated, including AV. \textbf{Full}: inter-group attention is unrestricted.\\

\begin{table}[!h]
\centering
\caption{Segregation ablation across LM-level and MM-level masking strategies.}
\label{tab:seg_princip}

\begin{threeparttable}

\footnotesize 

\setlength{\tabcolsep}{7pt} 
\begin{tabular}{lcccccc}
\toprule
$\mathrm{M}_{\text{LM}}$ & $\mathrm{M}_{\text{MM}}$ 
& \multicolumn{2}{c}{SIMS} 
& \multicolumn{2}{c}{MOSEI} \\
\cmidrule(lr){3-4} \cmidrule(lr){5-6}
 &  & MAE $\downarrow$ & Corr $\uparrow$ & Acc2 $\uparrow$ & F1 $\uparrow$  \\
\midrule

\multirow{3}{*}{Constr.}  
  & Constr. & \textbf{35.7} & \textbf{73.2} & \textbf{87.79} & \textbf{87.78} \\
  & Full    & 36.5 & 72.1 & 86.22 & 86.23 \\
  & Sym.    & 36.0 & 72.8 & 87.39 & 87.42 \\

\midrule

\multirow{3}{*}{Sym.}     
  & Constr. & 36.1 & 72.5 & 87.23 & 87.21 \\
  & Sym. & 36.3 & 72.3 & 87.12 & 87.12 \\
  & Full & 36.4 & 72.1 & 86.21 & 86.22 \\

\midrule

Full     
  & Full & 38.3 & 68.9 & 86.11 & 86.12 \\

\bottomrule
\end{tabular}

\end{threeparttable}
\end{table}

\autoref{tab:seg_princip} supports both claims consistently across datasets. Full/Full produces the largest degradation on both SIMS and MOSEI, with the most pronounced instability, confirming that unconstrained mixing during representation learning causes the cross-modal contamination the framework is designed to prevent. Replacing Full with Symmetric at the LM level substantially recovers performance on both benchmarks, clean unimodal pathways matter, but falls short of Constrained in all conditions. This gap isolates the contribution of the AV pathway's read-and-aggregate role: under Symmetric isolation, AV tokens can no longer read from evolving unimodal pathways, severing the accumulation of their layer-wise cross-modal history regardless of what MM-level masking is applied.
Within the proposed LM-level topology, varying the MM-level mask reveals a consistent pattern across both datasets. The Full variant incurs the largest drop within the Constrained-LM group, exposing the cost of the write direction specifically: allowing unimodal tokens to attend back to AV tokens during IGSA consolidation partially contaminates the clean pathways the LM-level mask preserved. This degradation is pronounced on MOSEI (-1.57 Acc2) and clearly visible on SIMS (+0.8 MAE), yet both remain well above Full/Full, confirming that LM-level segregation is the dominant contributor. The Symmetric MM variant partially recovers but consistently falls short of Constrained, confirming that AV tokens must retain post-injection read access to unimodal states to continue aggregating them after each refinement step.
The asymmetric topology of mutual isolation among unimodal groups with one-way read access for AV tokens emerges as the most effective structure among those evaluated; notably, this advantage holds across both the balanced modality contributions of SIMS and the text-dominant regime of MOSEI.

\subsubsection{Gate Mechanistic Validation}
\begin{figure}[!h]
    \centering
    \includegraphics[width=\columnwidth]{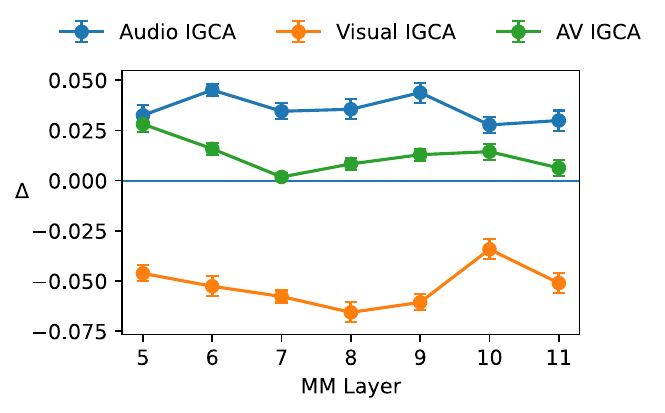}
    \caption{Gate activation shifts after removing the visual modality across the full SIMS test set. Visual gates close while audio gates compensate.}
    \label{fig:corrupt}
\end{figure}

To verify that the gates perform meaningful modality-specific regulation rather than uniform scaling, using SIMS, we remove the visual modality at inference and analyze changes in gate activations across layers (\autoref{fig:corrupt}). Three consistent patterns emerge: visual cross-attention gates close (negative $\Delta$), suppressing the missing stream; audio cross-attention gates open (positive $\Delta$), increasing reliance on the intact modality; and self-attention gates change minimally, indicating that reweighting occurs primarily at modality injection rather than intra-group refinement. Although not explicitly supervised, this behavior emerges from the element-wise gating design and demonstrates genuine modality-specific flow control. The effect is most pronounced in earlier MM layers, suggesting that modality reweighting is established during the initial injection stages and stabilizes as representations mature.

\section{Discussion}
\label{discuss}
SeRIn demonstrates that interaction topology is a meaningful design axis, distinct from depth and capacity. A component's utility cannot be evaluated independently of the structural context it operates within: the non-monotonic IGSA result and the capacity-matched Step 6 ablation both confirm that topology is the relevant unit of novelty, not the individual modules it organizes.
\noindent\textbf{Generalizability.} Fixing backbone, depth, and token count deliberately isolates topology as the sole variable. The Segregate principle is compatible with encoder-based architectures such as MAG-BERT and MulT: MulT's directional streams (A→T, V→T) already define distinct modality pathways, to which constrained inter-stream masking could apply directly to prevent premature mixing, without added modules or parameters. In early fusion models, the absence of structurally distinct modality pathways would require auxiliary objectives or gated bottlenecks, reintroducing the optimization-based specialization SeRIn is designed to replace. More broadly, we expect topology constraints to yield measurable gains when the architecture affords explicit modality pathways; verifying this hypothesis across other substrates and tasks beyond MSA is left as future work. A further open question is how the gates behave under weakly correlated or conflicting modality signals — for instance, divergent text and audio polarity — the regime where modality reweighting is most consequential; we leave this targeted analysis to subsequent work.
\noindent\textbf{SIMS vs. MOSEI.} The larger gains on SIMS are consistent with its near-balanced modality contributions: prior work documents that non-linguistic signals carry less weight in text-dominant settings~\cite{deepmlf}, so mechanisms protecting those signals plausibly have greater impact when modalities contribute more equally, though backbone scale, language, and dataset size are confounding factors. The consistent improvement across both datasets confirms that structured topology is 
beneficial under both regimes.




\section{Generative AI Use Disclosure}
Generative AI tools were used to assist with editing and polishing the language and presentation of this manuscript. All scientific content, experimental design, results, and conclusions are entirely the work of the authors. All co-authors have reviewed the final manuscript, take full responsibility for its content, and consent to its submission.



\end{document}